\documentclass[journal]{IEEEtran}

\usepackage{amsmath,graphicx,amssymb,multirow}
\usepackage{subfig}
\usepackage{epsfig}
\usepackage{tabularx,booktabs}
 \usepackage{graphicx,verbatim}
 
 \newcolumntype{C}{>{\centering\arraybackslash}X} 

\def\R{\mathbb R}

\ifCLASSINFOpdf

\else

\fi

\begin{document}

\title{SAR Image Despeckling Using a Convolutional Neural Network}

\author{Puyang~Wang, \emph{Student Member, IEEE}, He~Zhang,  \emph{Student Member, IEEE}\\ and Vishal M. Patel, \emph{Senior Member, IEEE}
\thanks{Puyang Wang, He Zhang, and Vishal M. Patel are with the Department
of Electrical and Computer Engineering, Rutgers University, Piscataway,
NJ, 08854 USA e-mail:\{puyang.wang, he.zhang92, vishal.m.patel\}@rutgers.edu}
}

\maketitle

\begin{abstract}
Synthetic Aperture Radar (SAR) images are often contaminated by a multiplicative noise known as speckle.  Speckle makes the processing and interpretation of SAR images difficult.  We propose a deep learning-based approach called, Image Despeckling Convolutional Neural Network (ID-CNN), for automatically removing speckle from the input noisy images. 
In particular, ID-CNN uses a set of convolutional layers along with batch normalization and rectified linear unit (ReLU) activation function and a component-wise division residual layer  to estimate speckle and it is trained in an end-to-end fashion using a combination of Euclidean loss and Total Variation (TV) loss.  Extensive experiments on synthetic and real SAR images show that the proposed method achieves significant improvements over the state-of-the-art speckle reduction methods.
\end{abstract}


\IEEEpeerreviewmaketitle

\section{Introduction}\label{sec:intro}
Synthetic Aperture Radar (SAR) is a coherent imaging technology that is capable of producing high resolution images of terrain and targets.   SAR has the ability to operate at night and in adverse whether conditions, hence it can overcome limitations of the optical and infrared systems.  However, SAR images are often contaminated by multiplication noise known as speckle \cite{Speckle_Goodman}.  Speckle is caused by the
constructive and destructive interference of the coherent returns scattered by small reflectors within each resolution cell.  The presence of speckle noise in SAR images can often make the processing and interpretation difficult for computer vision systems as well as human interpreters.  Hence, it is important to remove speckle from SAR images to improve the performance of various computer vision algorithms such as segmentation, detection and recognition.  

\begin{figure} [t!]
	\centering
	\subfloat[Speckled image]{%
		\includegraphics[width=0.45\linewidth]{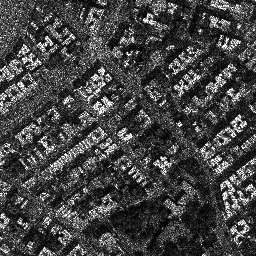}}
	\label{1a}\hspace{1em}
	\subfloat[Clean image]{%
		\includegraphics[width=0.45\linewidth]{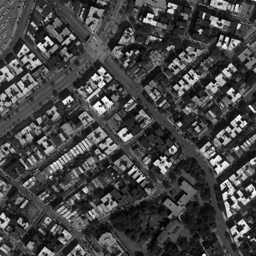}}
	\label{1b}\\
	
		\subfloat[Estimated noise]{%
			\includegraphics[width=0.45\linewidth]{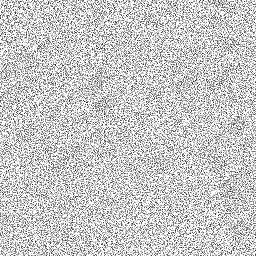}}
		\label{1a}\hspace{1em}
		\subfloat[Despeckled image]{%
			\includegraphics[width=0.45\linewidth]{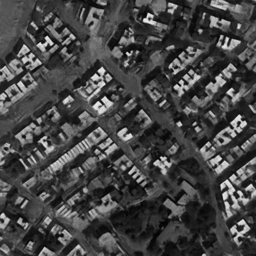}}
		\label{1b}
	\caption{A sample output of the proposed ID-CNN method for image despeckling.}
	\label{fig:sample_results} 
\end{figure}

Let $Y\in \R^{W\times H}$ be the observed image intensity, $X\in \R^{W\times H}$ be the noise free image, and $F\in \R^{W\times H}$ be the speckle noise.  Then assuming that the SAR image is an average of $L$ looks, the observed image $Y$ is related to $X$ by the following multiplicative model \cite{Book_Ulaby}
\begin{equation}\label{eq:multiplicative}
Y = F X,
\end{equation}
where $F$ is the normalized fading speckle noise random variable.  
One common assumption on $F$ is that it follows a Gamma distribution with unit mean and variance $\frac{1}{L}$ and has the following probability density function \cite{noisemodel}
\begin{align} \label{eq:pdf}
\centering
p(F) = \frac{1}{\Gamma(L)}L^LF^{L-1}e^{-LF},
\end{align}
where $\Gamma(\cdot)$ denotes the Gamma function and $F \geq 0$, $L \geq 1$.

Various methods have been developed in the literature to suppress speckle  including multi-look processing \cite{Book_SAR_IMU, Thompson_SAR}, filtering methods \cite{lee1981speckle, frost, gammamap}, wavelet-based despecking methods \cite{DeSpeckle_wavelet_MRF, Despeckle_Wavelet_undecimeted, Despeckle_wavelet_heavytail, Despeckle_MCA}, block-matching 3D (BM3D) algorithm \cite{bm3d} and Total Variation (TV) methods \cite{Despeckle_TV}.   
Note that some of these methods apply homomorphic processing in which the multiplicative noise is transformed into an additive noise by taking the logarithm of the observed data \cite{Despeckle_MCA}.  Furthermore, due to local processing nature of some of these methods, they often fail to preserve sharp features such as edges and often contain block artifacts in the denoised image.  

\begin{figure*}[t]
	\centering
	\includegraphics[width=170mm]{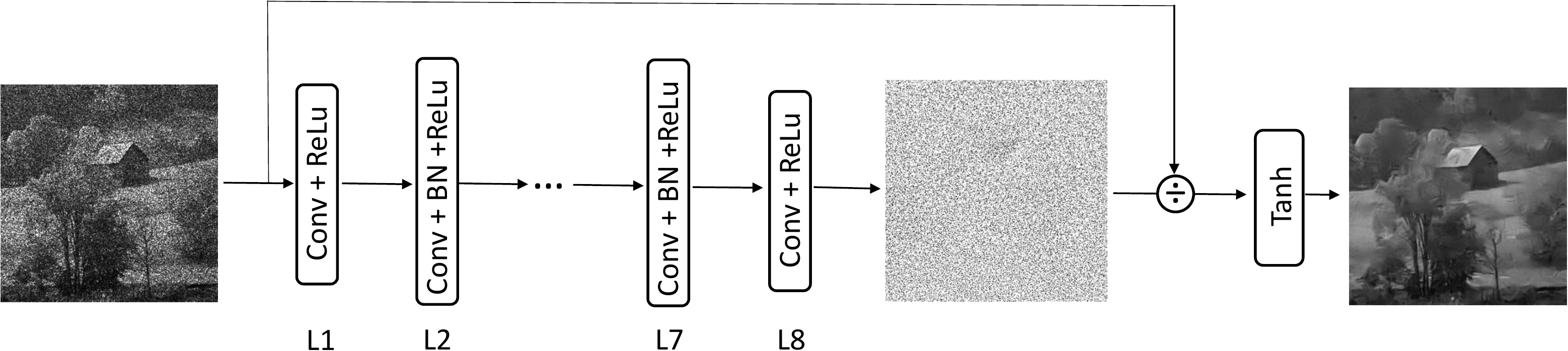}
	\caption{Proposed  ID-CNN network architecture for image despeckling.}
	\label{fig:network}
\end{figure*}

In recent years, deep learning-based methods have shown to produce state-of-the-art results on various image processing tasks such are image restoration \cite{cnn, cnnderain} and super resolution \cite{cnnsuper}.    However, to the best of our knowledge, image despeckling methods based on deep learning have not been extensively studied in the literature.   Motivated by recent advances in deep learning, we propose a Convolutional Neural Network (CNN) based approach for image despeckling.  Rather than using a homomorphic transformation \cite{Despeckle_MCA}, we directly estimate speckle using the input images based on the observation model \eqref{eq:multiplicative}.  The proposed  Image Despeckling Convolutional Neural Network (ID-CNN) method consists of several convolutional layers along with batch normalization \cite{batch_normalization} and rectified linear unit (ReLU) \cite{imagenet} activation function (see Figure~\ref{fig:network}).  In particular, the proposed architecture features a component-wise division-residual layer with skip-connection to estimate the denoised image.   It is trained in an end-to-end fashion using a combination of Euclidean loss and Total Variation (TV) loss.  One of the main advantages of using deep learning-based methods for image despeckling is that they learn parameters for image restoration directly from the training data rather than relying on pre-defined image priors or filters.
Figure~\ref{fig:sample_results} shows a sample output from our ID-CNN method.  Given the noisy image in Figure~\ref{fig:sample_results} (a), ID-CNN estimates the denoised image shown in Figure~\ref{fig:sample_results} (d).  The estimated speckle is also shown in Figure~\ref{fig:sample_results} (c).  As can be seen by comparing Figure~\ref{fig:sample_results} (b) and (d), one can see that out method is able to denoise the speckled image reasonably well. 


%

\section{Proposed Method}\label{sec:method}
In this section, we provide details of the proposed ID-CNN method for image despeckling. The detailed architecture of the proposed ID-CNN is shown in Figure~\ref{fig:network} where Conv, BN and ReLu stand for Convolution and Batch Normalization and Rectified Linear Unit, respectively.  Using a specific CNN architecture, we learn a mapping from an input SAR image to a despeckled image.  One possible solution to this problem would be to transform the image into a logarithm space and then learn the corresponding mapping via CNN \cite{sarcnn}. However, this approach needs extra steps to transfer the image into a logarithm space and from a logarithm space back to an image space. As a result, the overall algorithm can not be learned in an end-to-end fashion.  To address this issue, a division residual method is leveraged in our method where a noisy SAR image is viewed as a product of speckle with the underlying clean image (i.e. \eqref{eq:multiplicative}).  By incorporating the proposed component-wise division residual layer into the network,  the convolutional layers are forced to learn the speckle component during the training process.  In other words, the output before the division residual layer represents the estimated speckle.  Then, the despeckled image is obtained by simply dividing the input image by the estimated speckle.

\subsection{Network Architecture}
The noise-estimating part of ID-CNN network consists of eight convolutional layers (along with batch normalization and ReLU activation functions), with appropriate  zero-padding to make sure that the output of each layer shares the same dimension with that of the input image.  Each convolutional layer (except for the last convolutional layer) consists of 64 filters with stride of one. Then the division residual layer with skip connection divide the input image by the estimated speckle noise. A hyperbolic tangent layer is stacked at the end of the network which serves as a non-linear function. The details of the network configuration are given in Table~\ref{tab:network-config}. Here, L1 and L8 stand for the sequence of Conv-ReLU layers as depicted in Figure~\ref{fig:network}. Similarly, L2 to L7 denote Conv-BN-ReLU layers.

\begin{table}[htp!]
	\renewcommand{\arraystretch}{1.3}
	\caption{Network Configuration.}
	\label{tab:network-config}
	\centering
	\begin{tabular}{c|c|c|c|c}
		\hline
		\hline
		& Layer &   Filter Size & \#Filters&Output Size   \\
		\hline
 		L1&Conv &  $3\times 3 \times 1$ & 64&$256 \times 256 \times 64$   \\
 		&ReLU &   & &$256 \times 256 \times 64$   \\
		\hline
 		&Conv &   $3\times 3 \times 64$ & 64 & $256 \times 256 \times 64$  \\
		L2-L7&BN &    & &$256 \times 256 \times 64$  \\
		&ReLU &    & &$256 \times 256 \times 64$  \\
		\hline
 		L8&Conv &  $3\times 3 \times 64$ & 1&$256 \times 256 \times 1$   \\
 		&ReLU &     & &$256 \times 256 \times 1$   \\
		\hline
		\hline
	\end{tabular}
\end{table}


\subsection{Loss Function}
Loss functions form an important and integral part of learning process, especially in CNN-based image reconstruction tasks. Several works have explored different loss functions and their combinations for effective learning of tasks such as super-resolution \cite{cnnsuper}, semantic segmentation \cite{semantic_segmentation}, and style transfer \cite{style_transfer}. Previous works on CNN-based image restoration optimized over pixel-wise L2-norm (Euclidean loss) or L1-norm between the predicted and ground truth images.

Given an image pair $\{X, Y\}$, where $Y$ is the noisy input image and $X$ is the corresponding ground truth, the per-pixel Euclidean loss is defined as

\begin{equation}
L_E(\phi_E) = \frac{1}{WH}\sum_{w=1}^{W}\sum_{h=1}^{H}\|\phi(Y^{w,h})-X^{w,h}\|_2^2,
\end{equation}
where $\phi$ is the learned network (parameters) for generating the despeckled output and 
\begin{align}
\hat{X} = \phi(Y^{w,h}).
\end{align}
Note that we have assumed that $X$ and $Y$ are of size $W \times H$.

While the Euclidean loss has shown to work well on many image restoration problems, it often results in various artifacts on the final estimated image.  To overcome this issue, an additional TV loss is incorporated into the loss function to encourage more smooth results.  The TV loss is defined as follows
\begin{multline}
L_{TV} =\\
 \sum_{w=1}^{W}\sum_{h=1}^{H}\sqrt{(\hat{X}^{w+1,h}-\hat{X}^{w,h})^2+(\hat{X}^{w,h+1}-\hat{X}^{w,h})^2}.
\end{multline}
Finally, the overall loss function is defined as follows
\begin{align}
L = L_E + \lambda_{TV} L_{TV},
\label{eq:loss}
\end{align}
where $L_E$ represents per-pixel Euclidean loss function and $L_{TV}$ is total variation loss. Here, $\lambda_{TV}>0$ is a pre-defined weight for the $TV$ loss function.  We adjust this parameter to control the importance of the $TV$ loss.  The overall loss function ensures that the recovered image contains pixel level details and at the same time maintains the smoothness.



\begin{figure*}[htp!]
 \centering
 \includegraphics[width=35mm]{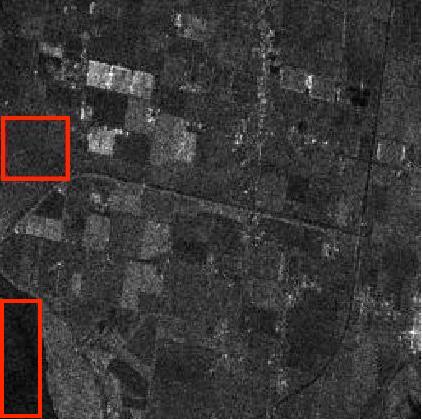}
 \includegraphics[width=35mm]{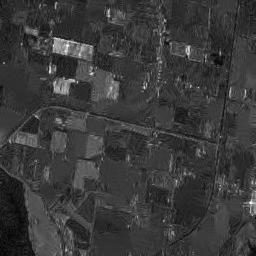}
\includegraphics[width=35mm]{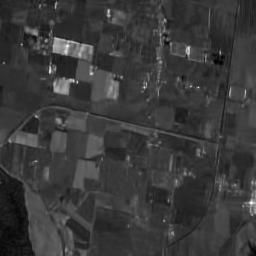}
\includegraphics[width=35mm]{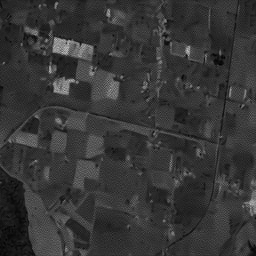}
\includegraphics[width=35mm]{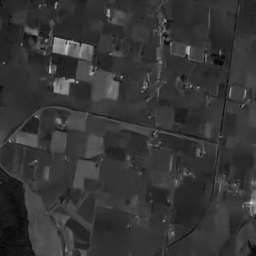}\\
\vspace{0.5em}
 \includegraphics[width=35mm]{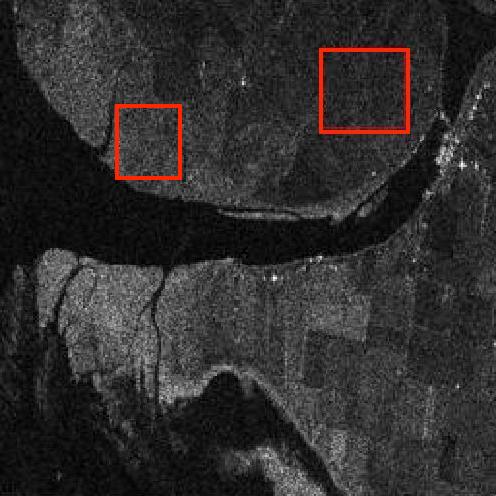}
 \includegraphics[width=35mm]{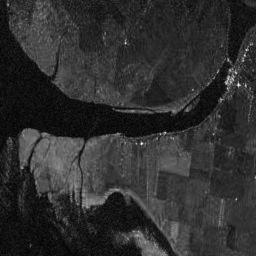}
\includegraphics[width=35mm]{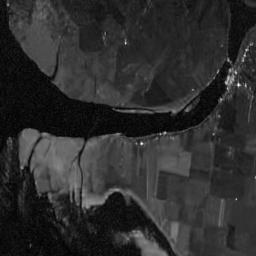}
\includegraphics[width=35mm]{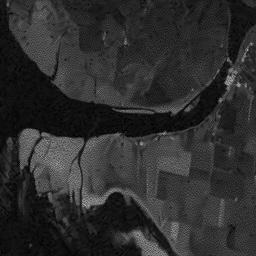}
\includegraphics[width=35mm]{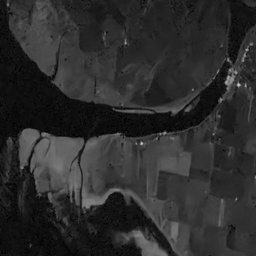}\\
 \caption{From left to right: SAR images, PPB, SAR-BM3D, SAR-CNN and ID-CNN.}
\label{fig:real}
\end{figure*}

\section{Experimental Results}\label{sec:results}
In  this  section,  we  present  the  results  of  our  proposed
ID-CNN algorithm on both synthetic and real SAR images.   We compare the performance of our method with that of the following six despeckling algorithms: Lee filter \cite{lee1981speckle}, Kuan filter \cite{kuan1985adaptive}, PPB \cite{ppb}, SAR-BM3D \cite{sarbm3d}, CNN \cite{cnn} and SAR-CNN \cite{sarcnn}.   Note that \cite{ppb, sarbm3d, cnn, sarcnn} are the most recent state-of-the-art image restoration algorithms.  For all the compared methods, parameters are set as suggested in their corresponding papers.  
For the basic CNN method, we adopt the network structure proposed in \cite{cnn} and train the  network using the same training dataset as used to train our network.   

\begin{figure}[htp!]
 \centering
 \includegraphics[width=85mm]{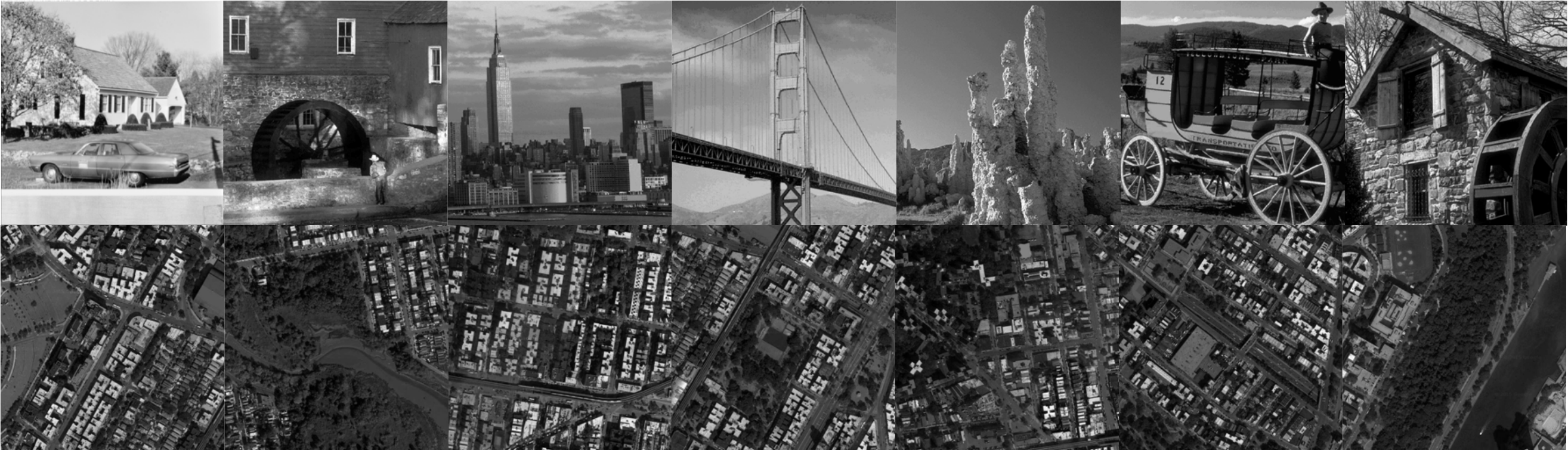}
 \caption{Sample images used to train our network.}
\label{fig:dataset}
\end{figure}

To train the proposed ID-CNN,  we generate a dataset that contains  3665 image pairs. Training images are collected from the UCID, BSD-500 and scraped Google Maps images \cite{pix2pix} and the corresponding speckled images are generated using \eqref{eq:multiplicative}.  All images are resized to $256 \times 256$. Sample images used for training our network are shown in Figure~\ref{fig:dataset}. The entire network is trained  using the ADAM optimization method  \cite{adam_opt}, with mini-batches of size 16 and learning rate of 0.0002. During training, the regularization parameter $\lambda_{TV}$ is set equal to $0.002$.  

\subsection{Ablation Study} 
\begin{figure} [htp!]
	\centering
	\subfloat[ID-CNN without TV loss]{%
		\includegraphics[width=0.45\linewidth]{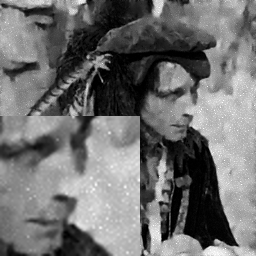}}
	\label{1a}\hspace{1em}
	\subfloat[ID-CNN with TV loss]{%
		\includegraphics[width=0.45\linewidth]{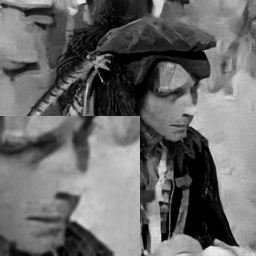}}
	\label{1b}
	\caption{Sample results of the proposed ID-CNN with and without TV loss. Note that white artifacts exist when total variation is not considered in the loss function.}
	\label{fig:TV} 
\end{figure}

We perform an ablation study to demonstrate the effects of different losses in the proposed method. Each of the losses are added one by one to the network and the results for each configuration is compared.  In the first configuration, only $L_E$ loss is minimized to train the network.  The number of looks, $L$, is set equal to 1.  The restored image is shown in Figure~\ref{fig:TV} (a).  It can seen that most of the speckle is removed from the noisy image, however, the denoised image still suffers from some artifacts.  When the network is trained by minimizing the overall loss which consists of both Euclidean and TV losses, the results get much better.  The use of TV loss removes those unwanted artifacts that were present in Figure~\ref{fig:TV}(a).   This can be clearly seen by comparing the zoomed in patches shown at the bottom left corner of these figures.  This experiment clearly shows the significance of having both Euclidean and TV losses in our framework. 

\subsection{Results on Synthetic Images}
We randomly selected 85 speckled images out of the 3665 images in the dataset.  The remaining 3580 images are used for training the network.  Experiments are carried out on three different noise levels. In particular, the number of looks $L$ is set equal to be 1, 4 and 10, respectively.  The Peak Signal to Noise Ratio (PSNR), Structural Similarity Index (SSIM) \cite{ssim}, Universal Quality Index (UQI) \cite{UQI} are used to measure the denoising performance of different methods.  Results corresponding this experiment are shown in Table~\ref{tab:synthetic-result}.  As can be seen from this table, in all three noise levels, ID-CNN provides the best performance compared to the other despeckling methods. Interestingly, the CNN method \cite{cnn} that directly learns a mapping from a noisy input image to a clean target image using a Euclidean loss performs worse than our method and PPB \cite{ppb} in many cases. This experiment clearly shows the significance of the proposed component-wise division residual layer-based CNN as well as the use of Euclidean + TV loss for image despeckling.

\begin{table*}[htp!]
	\renewcommand{\arraystretch}{1.3}
	\caption{Quantitative results for various experiments on synthetic images.}
	\label{tab:synthetic-result}
	\centering
	\begin{tabular}{c|c|c|c|c|c|c|c|c|c}
		\hline
		\hline
		&Metric  &  Noisy & Lee & Kuan & PPB & SAR-BM3D & CNN & SAR-CNN &ID-CNN \\
		\hline
		
		&PSNR   &  14.53 & 21.48 & 21.95 & 21.74 & 22.99 & 21.04 & 23.59 & \textbf{24.74}\\
		
		$L=1$&SSIM    &  0.369 & 0.511 & 0.592 & 0.619 & 0.692 & 0.630 & 0.640 & \textbf{0.727}\\
		
		&UQI       & 0.374 & 0.450 & 0.543 & 0.488 & 0.591 & 0.560 & 0.561 & \textbf{0.621}\\
		\hline
		&PSNR   &  18.49 & 22.12 & 22.84 & 23.72 & 24.96 & 22.60 & 26.20 & \textbf{26.89}\\
		
		$L=4$&SSIM& 0.525 & 0.555 & 0.650 & 0.725 & 0.782 & 0.722 & 0.771 & \textbf{0.818}\\
		
		&UQI    & 0.527 & 0.485 & 0.594 & 0.605 & 0.679 & 0.648 & 0.688 & \textbf{0.723}\\
		\hline
		&PSNR   &  20.54 & 22.30 & 23.11 & 24.92 & 26.45 & 23.52 & 27.63 & \textbf{28.07}\\
		
		$L=10$&SSIM&0.602 & 0.571 & 0.671 & 0.779 & 0.834 & 0.741 & 0.825 & \textbf{0.853}\\
		
		&UQI    & 0.599 & 0.498 & 0.613 & 0.678 & 0.745 & 0.683 & 0.741 & \textbf{0.765}\\
		\hline
		\hline
	\end{tabular}
\end{table*}

\begin{table}[htp!]
	\renewcommand{\arraystretch}{1.3}
	\caption{The estimated ENL results on real SAR images.}
	\label{tab:real-result}
	\centering
	\begin{tabular}{c|c|c|c|c|c}
		\hline
		\hline
		\# chip & PPB & SAR-BM3D & CNN & SAR-CNN & ID-CNN \\
		\hline
		
		1 & 42.49 & 69.26 & 32.32 & 50.76 & \textbf{89.43}\\
		
		2 & 8.63 & 10.95 & 7.50 & 8.93 & \textbf{13.90} \\
		
		3 & 103.25 & 127.38 & 31.65 & 99.13 & \textbf{193.00}\\
		
		4 & 34.84 & 63.83 & 7.65 & 43.13 & \textbf{69.40} \\

		\hline
		\hline
	\end{tabular}
\end{table}
\subsection{Results on Real SAR Images}
We also evaluated the performance of the proposed method and recent state-of-the-art methods on real SAR  images \cite{Book_Cumming}.   Since the true reflectivity fields are not available, we use the Equivalent Number of Looks (ENL) \cite{ENL} to measure the performance of different image despeckling methods.  The ENL values are estimated from the homogeneous regions (shown with red boxes in Figure~\ref{fig:real}).  The ENL results are also tabulated in Table~\ref{tab:real-result}. It can be observed from these results that the proposed  ID-CNN outperforms the others compared methods in all four homogeneous blocks. These results also demonstrate that our ID-CNN method can achieve much better performance in suppressing  speckle in real SAR images.

When a clean reference is missing, visual inspection is another way to qualitatively evaluate the performance of different methods. The despeckled results corresponding to the real images are shown in Figure~\ref{fig:real}.  The second to fifth columns of Figure~\ref{fig:real} show the despeckled images corresponding to PPB, SAR-BM3D, SAR-CNN and ID-CNN,  respectively. It can be observed that ENL results are consistent with the visual results. No obvious speckle exist in ID-CNN while PPB and SAR-CNN suffer from some noticeable
artifacts. It is also evident from these figures that filter-based reconstructions such as PPB and SAR-BM3D generally generate blurry edges compared to SAR-CNN and ID-CNN.

\subsection{Runtime Comparisons}
Experiments were carried out in Matlab R2016b using the MatConvNet toolbox, with an Intel Xeon CPU at 3.00GHz and an Nvidia Titan-X GPU. For ID-CNN, the entire network was trained using the torch framework \cite{torch}. Interestingly, once the training is over, ID-CNN exhibits the lowest run-time complexity as shown in Table~\ref{tab:runtime}. Note that even compared with the other CNN-based methods, our ID-CNN still has a lower run-time possibly because it has only eight fully convolutional layers compared to seventeen in other CNN-based methods \cite{cnn}, \cite{sarcnn}.

\begin{table}[htp!]
	\renewcommand{\arraystretch}{1.3}
	\caption{Runtime comparisons for despeckling an image of size 256 $\times$ 256.}
	\centering
	\begin{tabular}{c|c|c|c|c}
		\hline
		\hline
		PPB & SAR-BM3D & CNN & SAR-CNN &ID-CNN \\
		\hline
		
		9.82 s & 14.40 s & 2.42 s & 2.49 s & \textbf{0.56 s}\\
		
		\hline
		\hline
	\end{tabular}
	\label{tab:runtime}
\end{table}

\section{Conclusion}\label{sec:con}
We have proposed a new method of speckle reduction in SAR imagery based on CNNs. Compared to nonlocal filtering and BM3D image despeckling methods, our CNN-based method generates the despeckled version of a SAR image through a single feedforward process. Another uniqueness of our method is that the network is designed to recover the noise by convolutional layers and then the noisy input is divided by the estimated noise which results in the denoised output. This strategy, similar to residual learning, is inspired by the observation that a SAR image can be viewed as a product of a clean reference and noise. Results on synthetic and real SAR data show promising qualitative and quantitative results. This new process is also valuable for many SAR image understanding tasks such as road detection, railway detection, ship wake detection, texture segmentation for agricultural scenes and coastline detection.

\section*{Acknowledgment}
This work was supported by an ARO grant W911NF-16-1-0126

\bibliographystyle{IEEEtran}
\bibliography{despeckle}

\end{document}